\documentclass[10pt,twocolumn,letterpaper]{article}

\usepackage[pagenumbers]{cvpr} 

\usepackage{graphicx}
\usepackage{amsmath}
\usepackage{amssymb}
\usepackage{booktabs}

\usepackage[pagebackref,breaklinks,colorlinks]{hyperref}

\usepackage[export]{adjustbox}

\usepackage[capitalize]{cleveref}
\crefname{section}{Sec.}{Secs.}
\Crefname{section}{Section}{Sections}
\Crefname{table}{Table}{Tables}
\crefname{table}{Tab.}{Tabs.}

\def\cvprPaperID{33} 
\def\confName{CVPR}
\def\confYear{2023}

\begin{document}

\title{Neural Sign Reenactor: Deep Photorealistic Sign Language Retargeting}

\author{Christina O. Tze \textsuperscript{1} \thinspace\quad Panagiotis P. Filntisis \textsuperscript{1} \thinspace\quad Athanasia – Lida Dimou\textsuperscript{4} \\ \thinspace\quad Anastasios Roussos \textsuperscript{2,3} \thinspace\quad Petros Maragos\textsuperscript{1}  \\
\\
\textsuperscript{1}School of Electrical \& Computer Engineering, National Technical University of Athens, Greece\\
\textsuperscript{2}Institute of Computer Science (ICS), Foundation for Research \& Technology - Hellas (FORTH), Greece
\\
\textsuperscript{3}College of Engineering, Mathematics and Physical Sciences, University of Exeter, UK
\\
\textsuperscript{4}Institute for Language and Speech Processing, Athena R.C., Greece
}


\maketitle

\begin{abstract}
   In this paper, we introduce a neural rendering pipeline for transferring the facial expressions, head pose, and body movements of one person in a source video to another in a target video. We apply our method to the challenging case of Sign Language videos: given a source video of a sign language user, we can faithfully transfer the performed manual (\eg handshape, palm orientation, movement, location) and non-manual (\eg eye gaze, facial expressions, mouth patterns, head, and body movements) signs to a target video in a photo-realistic manner. Our method can be used for Sign Language Anonymization, Sign Language Production (synthesis module), as well as for reenacting other types of full body activities (dancing, acting performance, exercising, etc.). We conduct detailed qualitative and quantitative evaluations and comparisons, which demonstrate the particularly promising and realistic results that we obtain and the advantages of our method over existing approaches.
\end{abstract}

\vspace{-0.5cm}
\section{Introduction}
\label{sec:intro}
\vspace{-0.2cm}
One of the most challenging open problems of Sign Language (SL) technologies is the generation of synthetic SL videos that allow SL users to experience natural and fluid communication, similar to human-to-human communication. Prior to the deep learning era, the SL Production (SLP) problem was historically tackled using animated avatars
(\eg VisiCast \cite{bangham2000virtual}, Tessa \cite{cox2002tessa}, eSign \cite{zwitserlood2004synthetic} and Dicta-Sign \cite{efthimiou2012dicta}). However, in terms of the avatars' appearance and motion, this typically resulted in a low level of realism, reducing the plausibility and engagement of users with such technologies. 

\begin{figure}[h]
    \centering
    \adjincludegraphics[width=.8\columnwidth, trim={0 0 {.5\width} 0}, clip]{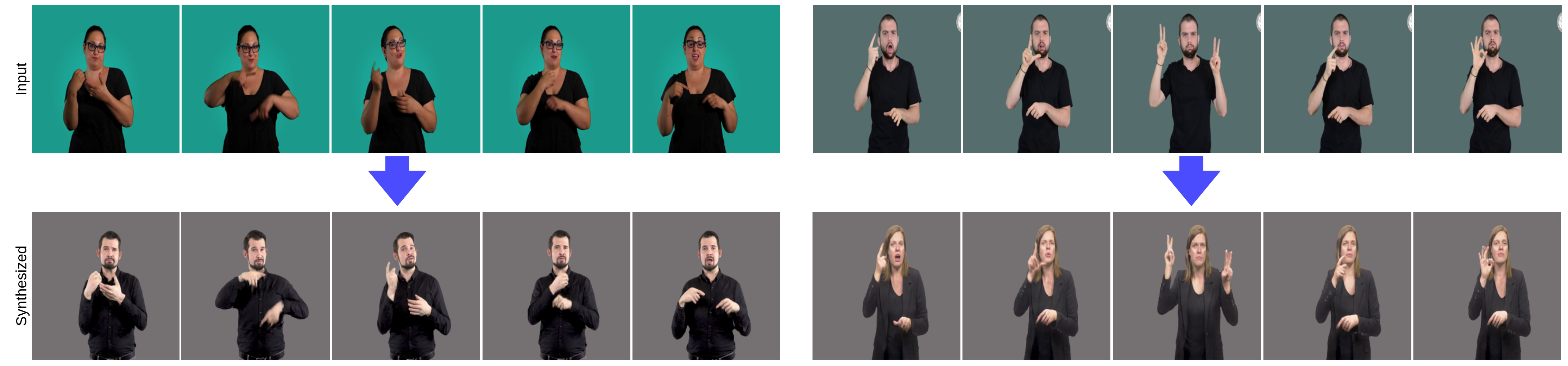}
    \vspace{-0.35cm}
    \caption{Given an input sign language video, our \textit{Neural Sign Reenactor} synthesizes a photo-realistic and temporally coherent video of a target signer imitating the source signer's body and facial movements. Please also refer to Suppl. Video \cite{suppvideo}.}
    \vspace{-0.7cm}
    \label{fig:teaser}
\end{figure}

With the advent of deep learning, novel methods have been introduced that build upon the latest advances in photo-realistic neural rendering and synthesize SL videos with avatars that have the appearance of real persons. Initial approaches (\eg \cite{stoll2018sign, stoll2020text2sign, zelinka2020neural}) dealt with this problem by concatenating isolated signs disregarding the natural co-articulation between them. In addition, other works (\eg \cite{zelinka2020neural, saunders2020progressive, saunders2020adversarial}) used skeleton pose representations rather than photo-realistic videos, which was shown to reduce Deaf understanding \cite{ventura2020can}. To improve sign comprehension, more recent approaches go one step further and apply human motion retargeting techniques to transform the predicted skeleton pose sequences into a photo-realistic human actor video. Human motion retargeting is an emerging topic at the intersection of computer vision and graphics due to its extensive potential for content creation. Over the last years, a plethora of deep learning-based methods has been introduced in this field. Some of them require high-fidelity 3D pose estimation or reconstruction \cite{liu2019neural,villegas2018neural, liu2019liquid,lim2019pmnet}. Retargeting motion from 2D inputs has also been studied in several works \cite{chan2019everybody,aberman2019deep,aberman2019learning,yang2020transmomo,zhu2022mocanet}. 
SignGAN \cite{saunders2020everybody} was the first SLP model to produce photo-realistic continuous SL videos by conditioning synthesis on the predicted skeletal pose sequence and the style image of a reference signer. The closest work to this paper is that of Saunders \etal~\cite{saunders2021anonysign}, who presented a deep learning framework for the generation of photo-realistic retargeted videos, using novel synthesized human appearances instead of the original signer appearance. However, their generated frames include artifacts and the synthesized human appearances are not always convincing as being real. This work overcomes the aforementioned limitations and synthesizes videos of unprecedented realism that include the upper body movements and facial expressions of a virtual signer who is almost indistinguishable from a real person. The motivation of our work is discussed in further detail in the Suppl. Material. Our contributions can be summarized as follows: 
\textbf{1)} We build upon an effective combination of two different body trackers for implementing high-fidelity body and face tracking.
\textbf{2)} We propose a novel scheme for conditioning the neural renderer.
\textbf{3)} We introduce a novel pose retargeting step that enables our model to work reliably across signers of different genders and body structures.
\textbf{4)} We conduct detailed qualitative and quantitative evaluations and user studies to evaluate our method and compare it with previous human motion transfer methods. The experiments demonstrate the particularly promising and realistic results that we obtain under challenging continuous signing scenarios.


\begin{figure*}
    \centering
    \includegraphics[width=.75\textwidth]{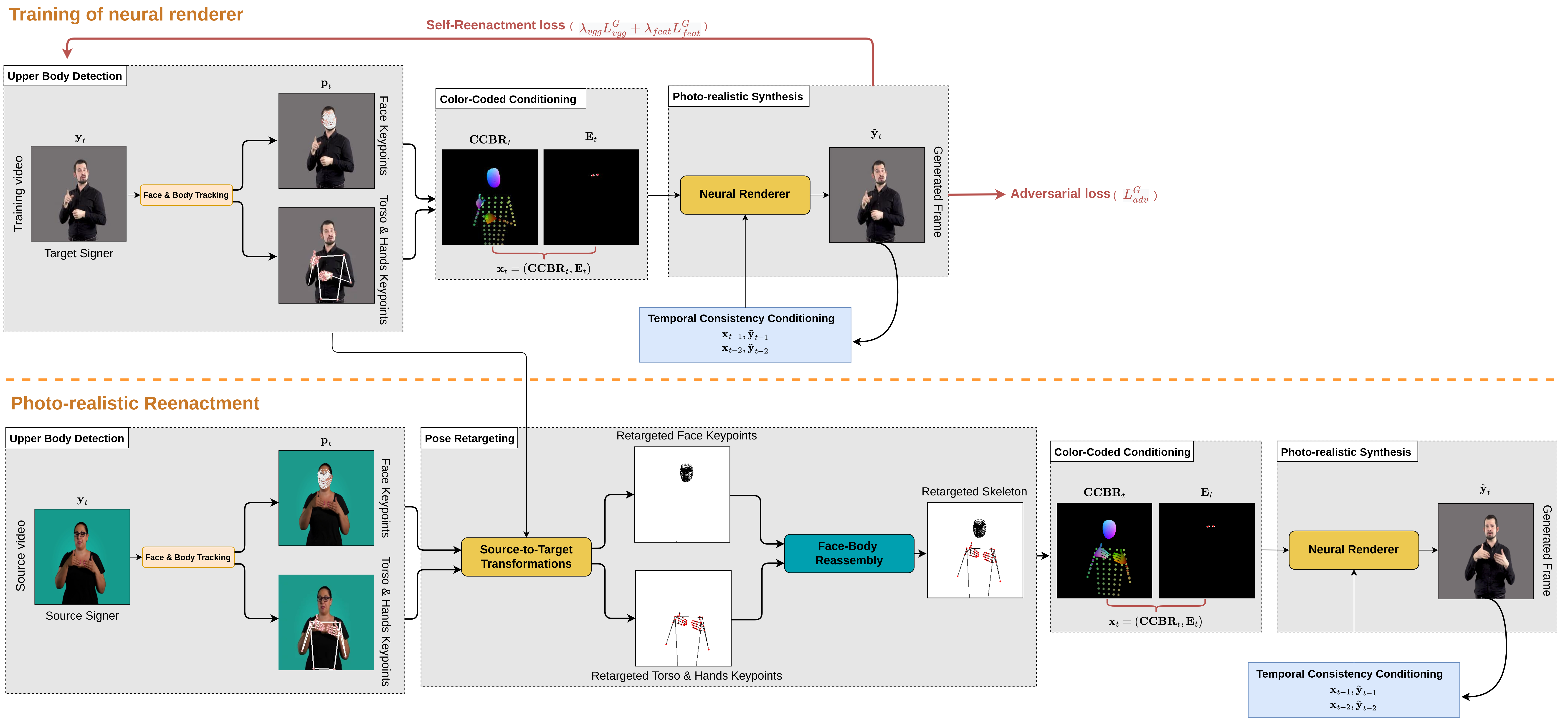}
    \vspace{-0.4cm}
    \caption{(Top) \textbf{Training}: We extract each target signer's skeleton pose sequence from his/her training video and use it to generate the corresponding color-coded body representations and eye gaze images, which are concatenated and fed into the neural renderer as conditional input. 
    (Bottom) \textbf{Reenactment}: We extract the source signer's skeleton pose sequence from his/her source video and then transform the estimated landmarks to match the target actor's body shape and location within each frame. The output frames are generated by the neural renderer from the corresponding conditional inputs using the previously trained model for the specific target signer.}
    \vspace{-0.6cm}
    \label{fig:pipeline}
\end{figure*}

\vspace{-0.2cm}
\section{Methodology}
\vspace{-0.2cm}
Given an input video ${\textbf{Y}}$, our method generates a photo-realistic and temporally coherent video $\tilde{\textbf{Y}}$ of a target actor imitating the source actor's upper body movements and facial expressions. An overview of the proposed pipeline is presented in Fig. \ref{fig:pipeline}. It consists of four main components: \\ 

\vspace{-0.42cm}
\noindent


\noindent \textbf{1) Upper Body Detection:}
We first extract the skeleton pose sequences from the SL videos (source and target) using the MediaPipe (MP)~\cite{lugaresi2019mediapipe} Pose and Holistic modules. More specifically, we use MP Holistic to track the head and hands, inferring 520 landmarks in total, while for the torso we use 9 from the 33 3D landmarks detected by the MP Pose model (since the Holistic module does not capture the depth of the pose landmarks).
After preprocessing, every frame is represented by a pose vector that stores the 3D coordinates of $K = 529$ tracked joints. 
Hereafter, the joint in the middle of the shoulders will be referred to as the \textit{root} joint. Finally, for every video, we crop every frame with a fixed-size and fixed-position bounding box that surrounds all locations of the skeleton's joints over all frames, leaving at each side (left/right/top/bottom) a margin whose size is 5\% of the corresponding average dimension (width/height). The cropped frames are then resized to the constant resolution of {\bf{256$\times$256 pixels}}. \vspace{0.05cm}



\noindent
\textbf{2) Pose Retargeting:} We propose a novel pose retargeting algorithm based on \textbf{Procrustes Analysis} for transferring the motion from a source character to a target. It is applied separately for two parts of the upper body, namely the head and the torso along with the hands, taking into consideration possible
differences between their body shapes. For the \textbf{head}, similarly to \cite{beeler2014rigid}, we use a subset of $n=94$ facial landmarks from the most rigid area of the face that are less affected by facial deformations during facial expressions and mouth motions. 
For the sake of simplicity, we call these landmarks \textbf{rigid} and all the rest \textbf{non-rigid}. 
For every frame and every video (either source or target), we consider the set of rigid 3D landmarks and perform Procrustes Analysis \cite{umeyama1991least} to rigidly align them to a common reference face template, which is defined in an anatomical coordinate system with axes aligned to the axial, coronal and sagittal planes. For each video (either source or target), we consider the aligned rigid landmarks and apply \textbf{geometric median} \cite{vardi2000multivariate} over all frames to extract a median face that robustly approximates the subject's facial geometry.
To account for cross-subject anatomical differences in the facial shape, we find the non-uniform per-axis scaling $\mathcal{S}$ that optimally registers the median face of the source to the median face of the target. Note that both median faces live in the anatomical coordinate system, therefore considering per-axis scaling only provides a satisfactory approximation.  
Finally, for each frame of the source video, we
consider all facial landmarks (rigid and non-rigid) and apply the following transformations: $\mathcal{T}1$) the already estimated Procrustes transformation from the source domain to the anatomical coordinate system, $\mathcal{T}2$) the non-uniform scaling $\mathcal{S}$, $\mathcal{T}3$) the inverse of transformation $\mathcal{T}1$.
For the \textbf{remaining part} of the upper body, we follow a similar procedure to that outlined for the head, ending up with two independent skeletons: one for the target subject's head pose and the other for his/her torso and hands movements. However, since the final sequence of retargeted skeletons must match the target actor's upper body movements, additional translations are required to combine the two separate skeletons into one and then adjust its overall position. To achieve this, every head skeleton in the output sequence is first attached to the nose joint of the corresponding torso skeleton. As a final step, 
a global translation and scaling are applied to the unified skeleton (head, torso, hands) to align it with the target subject's median scale and position at the target video's domain. This helps the neural renderer during reenactment since it ensures that the retargeted skeleton is as similar to the skeleton of the training video as possible. 

\vspace{-0.161cm}
\noindent
\textbf{3) Color-coded Conditioning:} 
Having adjusted the motion of the source person subject to the body shape and location of the target person, we follow \cite{kim2018deep, doukas2021head2head++} and generate convenient for neural rendering semantic representations of the body pose in the 2D image space, which we term \textbf{color-coded body representations}, \textbf{CCBR} $\in \mathbb{R}^{256 \times 256 \times 3}$. In more detail, these representations are 8-bit RGB images where each tracked joint is plotted as a disk of fixed radius and assigned a unique color based on a novel coloring scheme of a template skeleton. Please refer to Suppl.~Material for more details. 
Moreover, we found out that increasing the number of skeleton joints boosted our reenactment performance, and therefore we apply bone interpolation as a data augmentation technique, where both the color and number of interpolated points along a certain bone are fixed. For their coloring, we interpolate between the RGB colors of the tracked joints that define each bone. Similarly to \cite{doukas2021head2head++}, we also condition our video rendering network to \textbf{eye gaze images}, $\textbf{E} \in \mathbb{R}^{256 \times 256 \times 3}$, which are generated by drawing the left and right pupils as disks of fixed radius and connecting the eyes' contour landmarks. At each time step $t$, the CCBR is concatenated with the corresponding eye gaze image and fed to the neural renderer as conditional input, $\textbf{x}_t = ({\bf{CCBR}}_{t},{\bf{E}}_{t}) \in \mathbb{R}^{256 \times 256 \times 6}$.

\vspace{0.05cm}
\noindent
\textbf{4) Photo-realistic Synthesis:} 
We build upon the publicly available video rendering network of Head2Head++ \cite{doukas2021head2head++} for producing photo-realistic, temporally coherent videos. Our neural renderer is person-specific, which indicates that it is trained separately for every target actor using his/her reference video as the only training data. During training, we follow a self-reenactment setting where the source signer coincides with the target, thus we have access to the ground truth frames. The network consists of: \textbf{a)} a \textit{Generator} $G$, \textbf{b)} an \textit{Image Discriminator} $D_{I}$, and \textbf{c)} a \textit{Dynamics Discriminator} $D_{D}$. In contrast to \cite{doukas2021head2head++}, we also use a body segmentation model to prevent some artifacts in the background of the generated images. In terms of the network's architecture and training process, we follow Head2Head++. Please refer to Suppl.~Material for more details.

\vspace{-0.24cm}
\section{Comparison with other methods}
\vspace{-0.2cm}
We compare our method with two previous human motion transfer methods, namely Everybody Dance Now (EDN) \cite{chan2019everybody} and Video-to-Video Synthesis (Vid2Vid) \cite{wang2018video}. It is important to note that these approaches have been tested for reenacting full body activities, but we were unable to find a method that addresses the same problem as us and also has source code available. For additional results and visualizations, please refer to Suppl. Video \cite{suppvideo}. 

\begin{figure*}[t]
    \centering
    \adjincludegraphics[width=.8\textwidth, trim={0 {.5\height} 0 0}, clip]{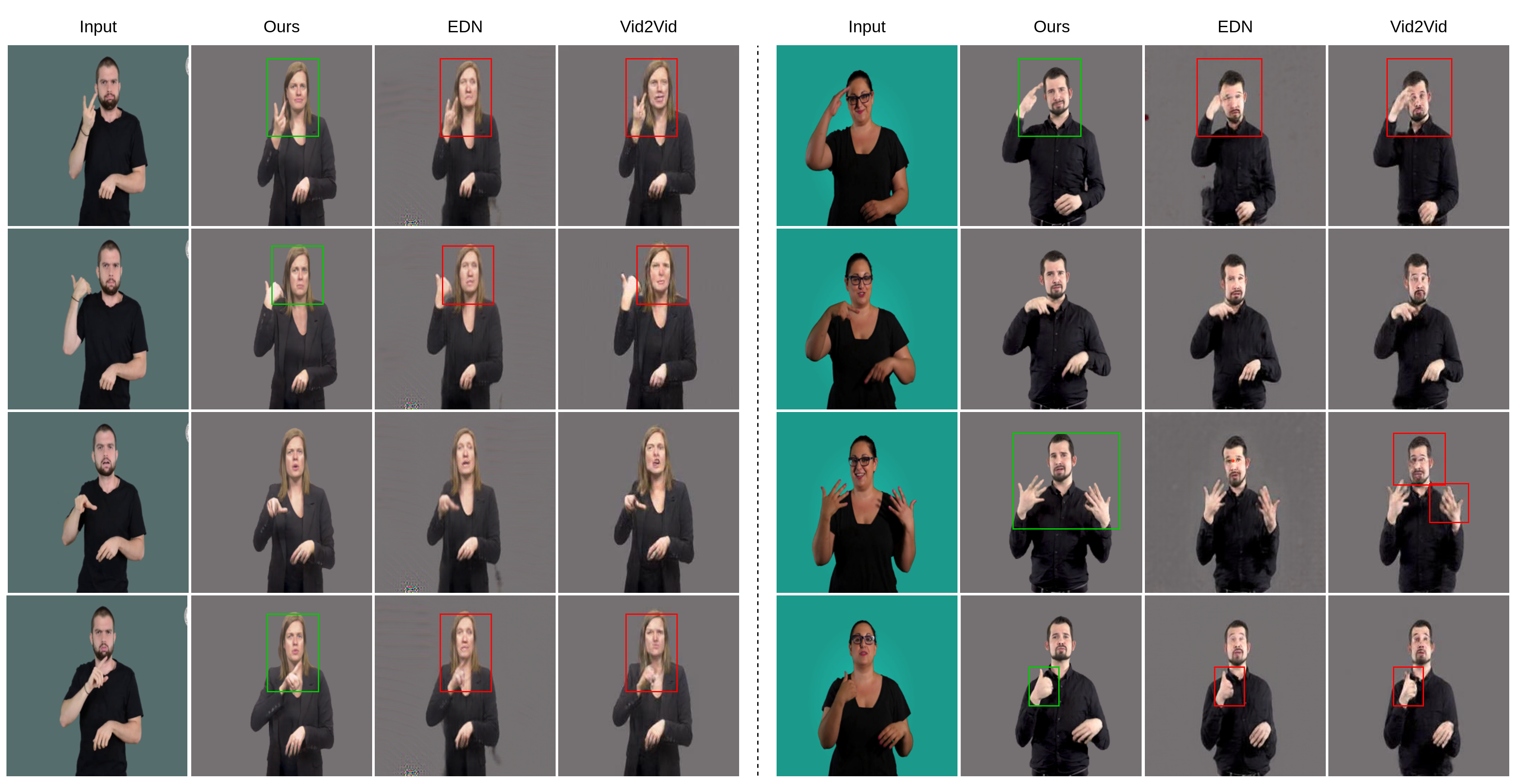}
    \vspace{-0.3cm}
    \caption{Visual comparison with EDN \cite{chan2019everybody} and Vid2Vid \cite{wang2018video} on different reenactment examples. We illustrate some erroneous results with red boxes and some successful examples of preserving the original mouth patterns and handshapes with green boxes. Please zoom in for details and refer to Suppl. Video \cite{suppvideo}.}
    \vspace{-0.6cm}
    \label{fig:qualitative}
\end{figure*}

\vspace{-0.5cm}
\paragraph{Qualitative Results:}
Fig.~\ref{fig:qualitative} displays the qualitative results of the three methods for a few representative frames of a male and female source actor signing in Greek SL (GSL). As can be seen, our method is capable of efficiently transferring the source person's head, torso, and hands movements, facial expressions, and eye gaze to the target subject. It also works reliably for different body types, generating frames with respect to the target subject's body structure. Moreover, it is evident that our approach outperforms the other two baselines in terms of both realism and pose transfer. In particular, we synthesize frames that look more realistic and natural, whereas EDN and Vid2Vid significantly distort the target's appearance. Compared to \cite{chan2019everybody} and \cite{wang2018video} that even use a specialized GAN to add realism
to a certain region (\eg Face GAN in EDN), we also result in a more accurate transfer of the source actor's facial expressions and handshapes to the target subjects. In general, our method generates photo-realistic videos of the target actor signing in the source actor's SL even though he/she has never used the particular SL and has never seen or performed the retargeted motions, which are determined by the input video.

\begin{figure}[t]
    \centering
    \adjincludegraphics[width=.8\columnwidth, trim={0 0 {.5\width} 0}, clip]{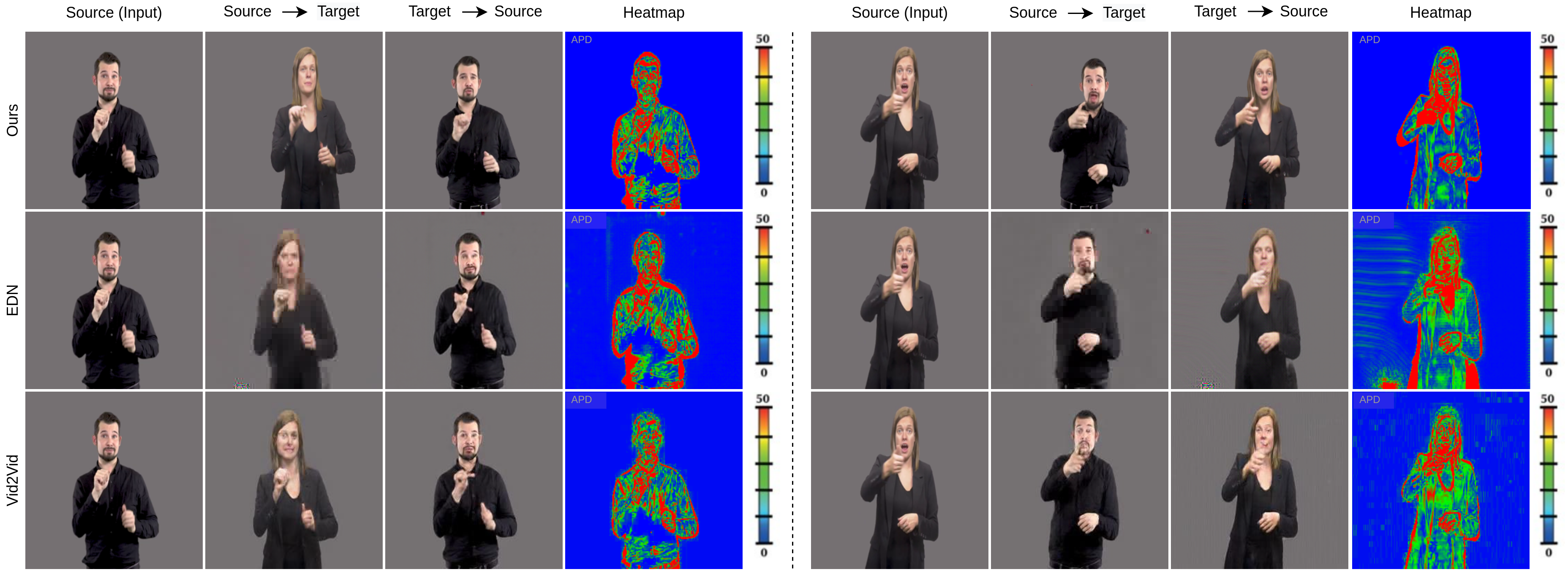}
    \vspace{-0.35cm}
    \caption{Cycle reenactment comparisons with EDN \cite{chan2019everybody} and Vid2Vid \cite{wang2018video}. From left to right: source actor, intermediate-target actor, original source actor driven by the manipulated target actor in the column before, and per-pixel differences between the first and third column.}
    \vspace{-0.6cm}
    \label{fig:cycle}
\end{figure}

\vspace{-0.66cm}
\paragraph{Quantitative Results:}
To assess the performance of the various methods, we conduct a \textbf{cycle reenactment} experiment, where the signing of a source actor is transferred to a target subject and then back to the same source. Ideally, the final video at the end of the experiment should be a reconstruction of the input one, so we can measure the per-pixel differences and calculate performance metrics. In particular, we use the \textbf{Average Pixel Distance (APD)} metric, which is computed as the average $l_{2}$ distance of RGB values across all pixels and frames, between the ground truth and final synthesized video. Table \ref{tab:cycle} shows the APD values for the three methods over the entire test sequence of our male and female target actors ($1,000$ frames each). As can be seen, our method outperforms EDN \cite{chan2019everybody} and Vid2Vid \cite{wang2018video} overall. Examples from our cycle reenactment experiments' results are displayed in Fig.~\ref{fig:cycle}. As already mentioned, our method synthesizes highly realistic frames, as opposed to the blurry and substantially distorted images that the other methods produce.
\vspace{-0.25cm}

\begin{table}[h!]
\scriptsize
\centering
\begin{tabular}{|c|c|c|c|} 
 \hline
 & Ours & EDN & Vid2Vid  \\ [0.5ex] 
 \hline\hline
 Male & 14.40 & 13.43 & \textbf{10.99}  \\ 
 \hline
 Female & \textbf{10.55} & 13.60 & 108.42  \\
 \hline
 Average & \textbf{12.48} & 13.52 & 59.71 \\
 \hline
\end{tabular}
\vspace{-0.35cm}
\newline\caption{Quantitative comparison of the three methods.}
\vspace{-0.6cm}
\label{tab:cycle}
\end{table}

\vspace{-0.3cm}
\paragraph{User Studies:}
\label{user}
We designed and implemented~\cite{kritsis2022danceconv} two user studies to evaluate the realism and faithful reenactment of different glosses from human users of GSL. 
\noindent The \textbf{first study} was a \textbf{Realism Study} which consisted of four questions, each including a pair of synthesized videos, one from our method and one from EDN or Vid2Vid, and asking the user to pick the one that seemed more realistic to him/her. The study was completed by 21 users and the preference results are presented in Table~\ref{tab:realism}. As can be seen, the overwhelming majority of users have rated our method as more realistic than the other two. \vspace{-0.25cm}

\begin{table}[h!]
\scriptsize
\centering
\begin{tabular}{c|c||c|c}
\multicolumn{2}{c||}{\textbf{Ours vs.~EDN}} & \multicolumn{2}{c}{\textbf{Ours vs.~Vid2Vid}} \\
\hline 
Ours & EDN & Ours & Vid2Vid \\
\hline
\textbf{(39/42) 92.9\%}   &  (3/42) 7.1\%  & \textbf{(40/42)  95.2\%} & (2/42) 4.8\%
\end{tabular}
   \vspace{-0.35cm}
   \newline\caption{Preference results on the realism of each method. Our method is \textbf{significantly} ($p \approx 10^{-9}$ and $p \approx 10^{-8}$, binomial test) more realistic compared to EDN and Vid2Vid.}
\label{tab:realism}
\vspace{-0.35cm}
\end{table}

\vspace{-0.13cm}
\noindent In our \textbf{second study}, which was a \textbf{Sign Classification Study}, we evaluated how faithfully each method reenacted a number of different GSL glosses. For that, we carefully selected based on the guidance of an SL expert 14 glosses and reenacted them using our method, EDN, and Vid2Vid. Then, we showed each user 12 glosses (3 for each method, plus 3 for the source videos) and asked them which gloss was being signed, from a list of 7 choices (including ``None of the above"). A total of 23 users completed this study, and the results are shown in Table~\ref{tab:recogn}, where we can see that all methods achieve high accuracy regardless of their realism, in the cost however of the user experience. The small discrepancies between the different methods are statistically insignificant and can be attributed to: \textbf{a)} the random sampling from the question bank leading to a slightly different distribution of glosses between the various methods and \textbf{b)} the fact that some participants might not have identified the specific signing style of the source actor for some glosses,  
leading them to mistakenly select ``None of the above" if the source video had a different signing style from the one they are familiar with. \vspace{-0.2cm}

\begin{table}[h!]
\scriptsize
\centering
\begin{tabular}{c|c|c|c}
Ours & EDN & Vid2Vid & Real video \\
\hline
(53/69) 76.8\% & \textbf{(55/69) 79.7\%} & (53/69) 76.8\% & (51/69) 73.9\% 
\end{tabular}
   \vspace{-0.35cm}
   \newline\caption{
   Classification accuracy of each method on different GSL glosses. 
   There is no significant difference between all methods ($p$=$1$ for all pairwise proportion tests with Bonferroni correction).}
\label{tab:recogn}
\vspace{-0.62cm}
\end{table}

\vspace{-0.03cm}
\section{Conclusions}
\vspace{-0.2cm}
We proposed \textit{Neural Sign Reenactor}, a novel neural rendering pipeline for transferring the body movements, head pose, and facial expressions of a source actor in a driving video to a target subject in a reference video. We have applied our approach to the challenging case of SL videos. Our extensive qualitative and quantitative evaluations have demonstrated that our method faithfully transfers the source signer's manual and non-manual signs to a target signer and works reliably across signers of different genders and body structures. Compared to earlier methods for human motion retargeting that dramatically alter the target subject's appearance, it produces highly realistic and natural-looking results. We believe that our work paves the way for the development of novel SLP systems that go beyond computer-generated avatars and produce photo-realistic SL videos increasing the appeal and engagement of the users.

\clearpage
{\small
\bibliographystyle{ieee_fullname}
\bibliography{main}
}

\clearpage

\section*{Supplementary Material}

\maketitle
\appendix

\section{Motivation}
In this section, we provide a more detailed discussion of the motivation of our work.

Tens of millions of Deaf worldwide use Sign Language (SL) as their native language \cite{who,bda,thinktank,antonakos2015survey}. At the same time, most of them have limited reading and writing skills in the spoken language, which for them is a foreign language with a fundamentally different grammatical structure. 
Because of that, the Deaf are still disadvantaged in many contexts of their daily life, such as social relations, education, work, usage of computers, and the Internet. SL technologies can be a valuable ally of the Deaf community in their struggle to overcome these barriers, by building systems that facilitate their communication with the rest population \cite{papastratis2021artificial}. This has been an active research area during the last three decades, but it was only in the last years that it started maturing, thanks to the introduction of novel deep learning methods that yielded highly robust and promising results on the challenging tasks of Sign Language Recognition (SLR) \cite{roussos2013dynamic, theodorakis2014dynamic, koller2019weakly, camgoz2020sign, zhou2020spatial, koller2020quantitative}, Translation (SLT) \cite{camgoz2018neural,yin2020better,voskou2021stochastic} and Production (SLP) \cite{stoll2020text2sign, saunders2020progressive, saunders2020adversarial}.

Deep learning approaches require the availability of large-scale SL corpora which is very limited due to participants' concerns over privacy and video misuse \cite{bragg2020exploring}. Therefore, there is an urge to increase the amount of publicly available data and thereby further improve the performance of SL systems. 
In addition, special attention must be paid to cases of videos of SL datasets that refer to third-party personal information (\eg~names or personal data of other people). At the same time, one of the important  barriers that the Deaf are currently facing is related to their ability for online participation, especially in cases where the option of anonymity is a valuable tool for constructing a safe space to discuss sensitive, controversial or personal topics in social media or other online platforms \cite{lee2021american}: In contrast to the users of spoken languages who can easily communicate anonymously by just typing a text, the SL users can only communicate in their native language by using a camera capturing their hands, body, and face during signing, which reveals their identity. Since all these body parts convey cues that are important for SL communication \cite{antonakos2015survey}, it becomes evident that there is no easy way to conceal the signers' identity through simple video editing approaches.   

The aforementioned problems have recently attracted the interest of the research community, resulting in some specialized systems that seek to anonymize SL videos. This is a particularly difficult task due to the challenges in capturing, representing, and retargeting the human motions during signing, for example: extremely fast motion and articulation of the hands, complex interactions between the different body parts (\eg~between the two hands or between each of the hands and the face), large variability and complexity of hand configurations, and inter-signer variations due to anatomical differences. Our method can conceal the identity of the original signers by reproducing their videos using other actors who have given their informed consent for their recordings to be shared publicly and therefore can support the Deaf in increasing their online participation.

Regarding the applications of our framework, it can also be beneficial for the following purposes: 1) It can be readily used as the backend module in SLP systems, offering the option to have virtual interpreters with the appearance of real persons,
going well beyond the traditional graphics-generated
avatars. 2) Although it is developed and
tested on the especially challenging problem of SL reenactment, it can be readily applied to other types of full
body activities (dancing, exercising, etc.).

\section{Color-coded Conditioning}
In this section, we provide more details about our novel color-coding scheme, which is used in the \textit{Color-coded Conditioning} module of our pipeline for generating the color-coded body representations and eye gaze images. 

\begin{figure}[h]
    \centering
    \adjincludegraphics[width=0.5\textwidth, clip]{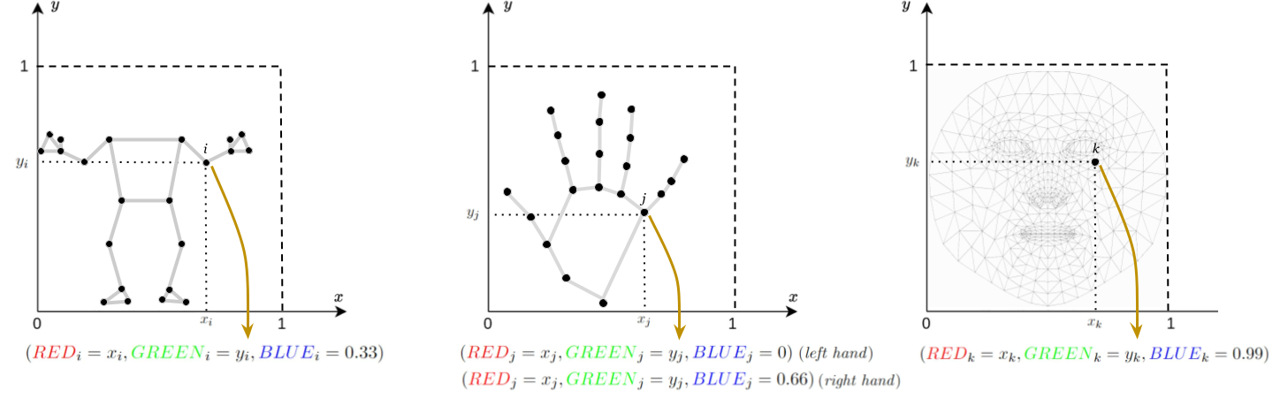}
    \vspace{-0.65cm}
    \caption{Visualization of our color-coding scheme for the torso, hands, and face.}
    \label{fig:color_coding}
\end{figure}

Regarding the \textbf{CCBRs}, the joints are colored using the following scheme, which assigns each joint a \textbf{unique color}: The $\textcolor{red}{Red}$ and $\textcolor{green}{Green}$ channels are given values directly from the $x$ and $y$ coordinates, respectively, of a template body's joints in the 2D image space, after being normalized between $0$ and $1$. Similarly, for the coloring of the face, we are based on its UV visualization and 2D texture coordinates by MediaPipe Face Mesh \cite{mediapipe_face}. The $\textcolor{blue}{Blue}$ channel has predefined and independent of the landmarks values for the torso, left hand, right hand, and face (see Fig. \ref{fig:color_coding}). Because we give each joint a unique, fixed color regardless of the signer, this indicates that all of them will have the exact same color in any such representation. This is why these representations are also referred to as semantic and they have generally been shown to help neural renderers learn the mapping to the output images since they are both in the RGB space. 

\begin{figure}[h]
    \centering
    \adjincludegraphics[width=\columnwidth, clip]{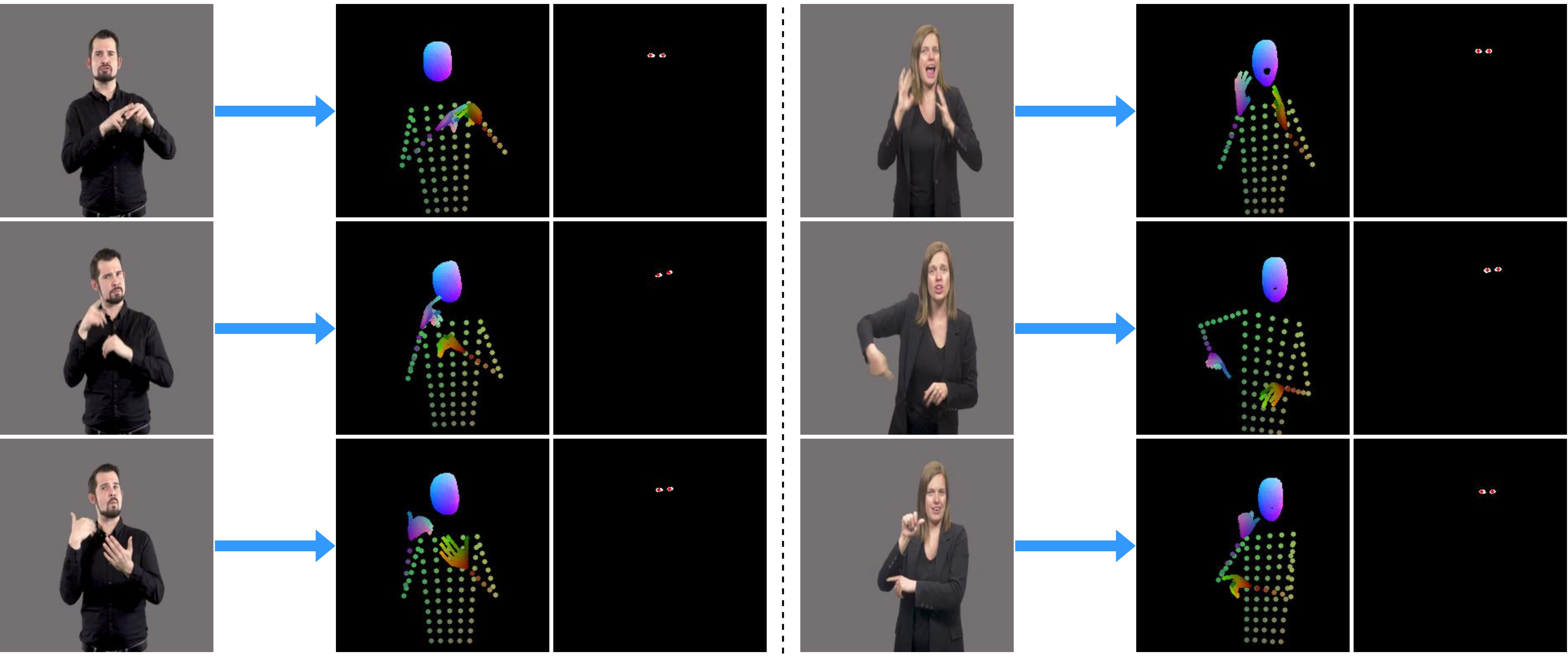}
    \vspace{-0.65cm}
    \caption{Examples of conditional inputs generation for some representative frames of the target
actors’ training videos. For each section, we illustrate from left to right: input frame, color-coded body representation, and eye gaze image. Please zoom in for details and refer to Suppl. Video \cite{suppvideo}.}
    \label{fig:conditional_inputs}
\end{figure}

As also mentioned in the main paper, in addition to the CCBRs, we condition our video rendering network to \textbf{eye gaze} images, which are generated by drawing the left and right pupils as disks of fixed radius and connecting the eyes' contour landmarks. For their coloring, we follow \cite{doukas2021head2head++} and tint the contour landmarks white and the pupils red. An illustrative example of all types of conditional inputs that we feed our neural renderer with is provided in Fig. \ref{fig:conditional_inputs}.

\section{Photo-realistic Synthesis}
As stated in the main paper, our video rendering network's components and training objectives are identical to Head2Head++ \cite{doukas2021head2head++}, thus they are briefly described below. 
\begin{itemize}
    \item \textbf{Generator} $G$: Given the conditional inputs $\textbf{x}_{t-2:t}$ of the current and the two preceding frames as well as the two previously generated frames $\tilde{\textbf{y}}_{t-2:t-1}$, the generator renders the frame of the output video at time step $t$:
\begin{align}
    \tilde{\textbf{y}}_{t} = G(\textbf{x}_{t-2:t}, \tilde{\textbf{y}}_{t-2:t-1})
\end{align}
The output video $\tilde{\textbf{Y}}_{1:T}$ shows the target subject performing the source signer's manual and non-manual signs, as determined by the conditional inputs sequence $\textbf{X}_{1:T}$. The Generator consists of two identical encoders, operating in parallel, as well as a decoder. The first encoder receives the conditional inputs $\textbf{x}_{t-2:t}$, while the second is given the two previously generated frames $\tilde{\textbf{y}}_{t-2:t-1}$. The two extracted feature maps are first added and then passed through the decoder, which brings the output $\tilde{\textbf{y}}_{t}$ in a normalised $[-1,+1]$ range, using a tanh activation function. 
\item \textbf{Image Discriminator} $D_{I}$: The image discriminator is used during training and aims at telling real and synthesized frames apart. At time step $t$, it receives the real pair $(\textbf{x}_{t}, \textbf{y}_{t})$ and the fake one $(\textbf{x}_{t}, \tilde{\textbf{y}}_{t})$.
\item \textbf{Dynamics Discriminator} $D_{D}$: The dynamics discriminator is used during training to enforce the temporal coherence of the output video. It receives a set of three consecutive real frames $\textbf{y}_{t:t+2}$ or fake frames $\tilde{\textbf{y}}_{t:t+2}$ along with the optical flow $\textbf{w}_{t:t+1}$, computed from the target's subject training video $\textbf{Y}_{1:T}$, and should learn to distinguish the fake data $(\textbf{w}_{t:t+1}, \tilde{\textbf{y}}_{t:t+2})$ from real data $(\textbf{w}_{t:t+1},\textbf{y}_{t:t+2})$. In this way, the generator tries to synthesize fake frames with the same flow/dynamics as the corresponding real ones in order to fool the discriminator.
\item \textbf{Objective function}: The total objective for $G$ consists of three terms:
\begin{align}
     L^{G} = L^{G}_{adv} + \lambda_{vgg}L^{G}_{vgg} + \lambda_{feat}L^{G}_{feat}
 \end{align}
 with $\lambda_{vgg} = \lambda_{feat} = 10$ as in \cite{doukas2021head2head++}.\\
 The first loss corresponds to the \textbf{adversarial objective} of the generator and is defined as in LSGAN \cite{mao2017least} using the 0-1 binary coding scheme ($b=c=1$ and $a=0$):
\begin{align}
     L^{G}_{adv} =  \frac{1}{2}\mathbb{E}_{t}[(D_{I}(\textbf{x}_{t}, \tilde{\textbf{y}}_{t})-1)^2] \nonumber\\ 
     +
     \frac{1}{2}\mathbb{E}_{t}[(D_{D}(\textbf{w}_{t:t+1}, \tilde{\textbf{y}}_{t:t+2})-1)^2] 
 \end{align}
 The second term is the \textbf{VGG loss}, which is computed as in \cite{wang2018high} and \cite{wang2018video}, by using the VGG network \cite{simonyan2014very} to extract feature representations in different layers for both the ground truth $\textbf{y}_{t}$ and the synthesized frame $\tilde{\textbf{y}}_{t}$ and then calculating their euclidean distance.\\
 The final loss in the generator's objective function is the overall \textbf{feature matching loss} which is equal to:
 \begin{align}
     L_{feat}^{G} =  L_{feat}^{G-D_{I}} +  L_{feat}^{G-D_{D}}
 \end{align}
 The first sub-loss, $L_{feat}^{G-D_{I}}$, is computed by extracting the activations on an intermediate layer of the image discriminator $D_{I}$ for a fake frame $\tilde{\textbf{y}}_{t}$ and the corresponding ground truth $\textbf{y}_{t}$ and then computing their $l_{2}$ squared distance. Similarly, $L_{feat}^{G-D_{D}}$ is computed using the Dynamics Discriminator $D_{D}$ instead of $D_{I}$.
\end{itemize}

\begin{figure*}[t]
    \centering
    \includegraphics[width=\textwidth]{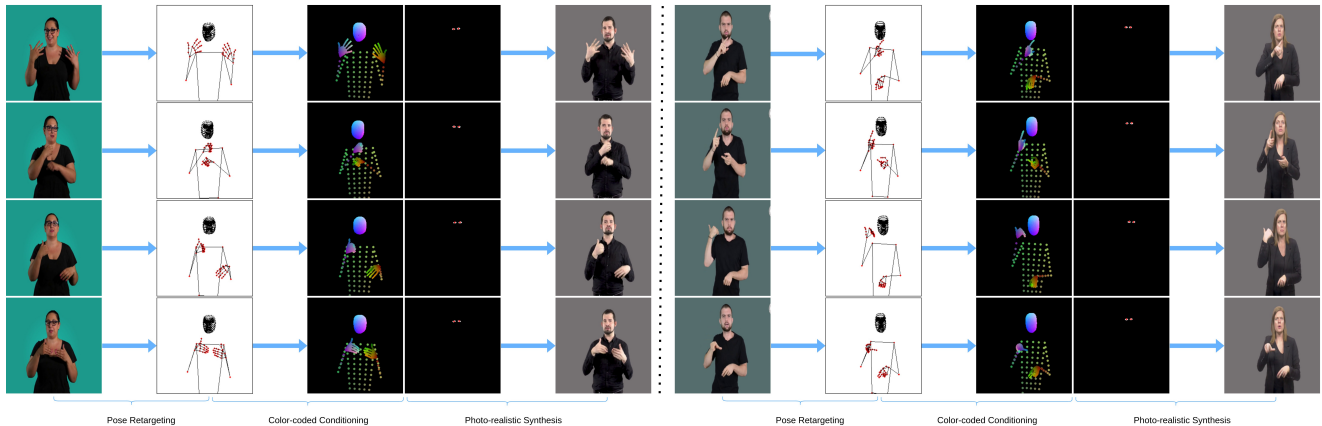}
    \vspace{-0.65cm}
    \caption{Visualization of intermediate and retargeted results for some representative frames of a female and male source actor from the Continuous Signing dataset. From left to right: input frame, retargeted skeleton, conditional inputs, and output frame.}
    \label{fig:additional_visualizations}
\end{figure*}

\section{Experimental Setup}
In this section, we describe the experimental setup of our method including the collected datasets and some implementation details. 
\subsection{Datasets}
We used \textbf{three datasets} for our experiments, which are presented below:\\
\\
\noindent \textbf{1)} \textbf{Target Actors dataset}: We selected 2 publicly available YouTube videos \cite{otc} for \textbf{training} our person-specific neural renderer. More specifically, we chose two individuals as our target subjects, a male and a female with different body types, signing in American Sign Language (ASL) and Quebec Sign Language (QSL), respectively. Each training video was at 30 fps and had approximately 10 minutes duration and $1280 \times 720$ spatial resolution. The frames of each subject were split into a training and a test set using a 90:10 split. It's crucial that the training videos show the target actors performing a wide range of upper body movements and facial expressions. \\ \\ 
\noindent \textbf{2)} \textbf{Source Actors dataset}: We collected a small dataset of 14 source videos from an online Greek Sign Language (GSL) dictionary \cite{dict1}, which we used to assess the performance of the various methods in our \textbf{Sign Classification Study}. Six individuals, four men and two women, were included in our source footage and each of them performed a distinct GSL sign that lasted from one to three seconds. Each actor's frames from this dataset were kept as test data and used for our reenactment experiments. In contrast to the training videos, we only require decent pose detection on the source footage. \\ \\
\noindent \textbf{3)} \textbf{Continuous Signing dataset}: We chose 4 publicly available videos \cite{omke,sdeng} of two male and two female actors signing continuously for $\approx$30 seconds each. Every video in this dataset was used as source footage and the performed signs were retargeted at the opposite gender's target subject, resulting in a total of four synthesized videos. These videos were included in our \textbf{Realism Study}. 

We’d like to clarify that our neural renderer is trained separately for every target actor and the only training data is his/her training video. Therefore, each trained model at the end is dedicated to a specific target subject from the training dataset, which is the Target Actors dataset. There is no need to train the neural renderer using the videos from the
remaining two datasets, i.e., Source Actors and Continuous Signing, because they only serve as source videos in our experiments.

\subsection{Implementation Details}
Our person-specific video rendering network requires a few minutes of footage for each target actor. In particular, for every subject in our Target Actors dataset, we used a $\approx$10-minute video and the training task (100 epochs) was completed in approximately 4 days on two NVIDIA GeForce GTX 1080 Ti GPUs. The networks were optimized using Adam \cite{kingma2014adam} with an initial learning rate $\eta = 2 \cdot 10^{-4}$, $\beta_{1} = 0.5$ and $\beta_{2} = 0.999$.

\section{Additional Visualizations}
\begin{figure}[h]
    \centering
    \adjincludegraphics[width=7cm, height = 5.3cm]{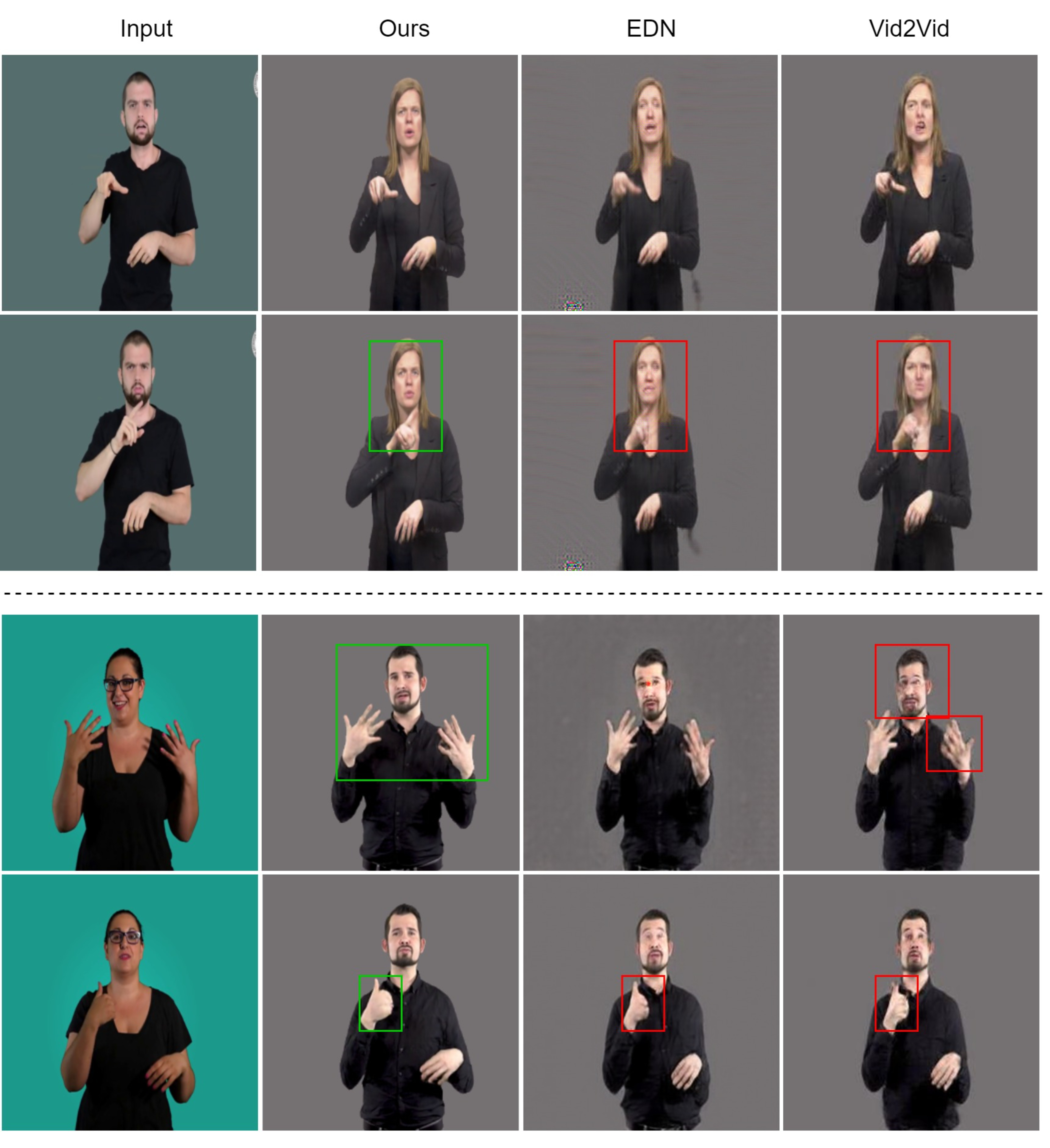}
    \vspace{-0.25cm}
    \caption{Visual comparison with EDN \cite{chan2019everybody} and Vid2Vid \cite{wang2018video} on different reenactment examples. We illustrate some erroneous results with red boxes and some successful examples of preserving the original mouth patterns and handshapes with green boxes. Please zoom in for details and refer to Suppl. Video \cite{suppvideo}.}
    \vspace{-0.2cm}
    \label{fig:qualitative_cropped}
\end{figure}

We show in Fig. \ref{fig:additional_visualizations} a more detailed view of our method's intermediate steps in a reenactment setting, including pose retargeting and color-coded conditioning. Also, Fig \ref{fig:qualitative_cropped} provides additional qualitative results of our method in comparison with EDN \cite{chan2019everybody} and Vid2Vid \cite{wang2018video} in the form of static frames for two actors of our Continuous Signing dataset. 

\begin{figure}[h]
    \centering
    \adjincludegraphics[width=6.5cm, height = 5.5cm]{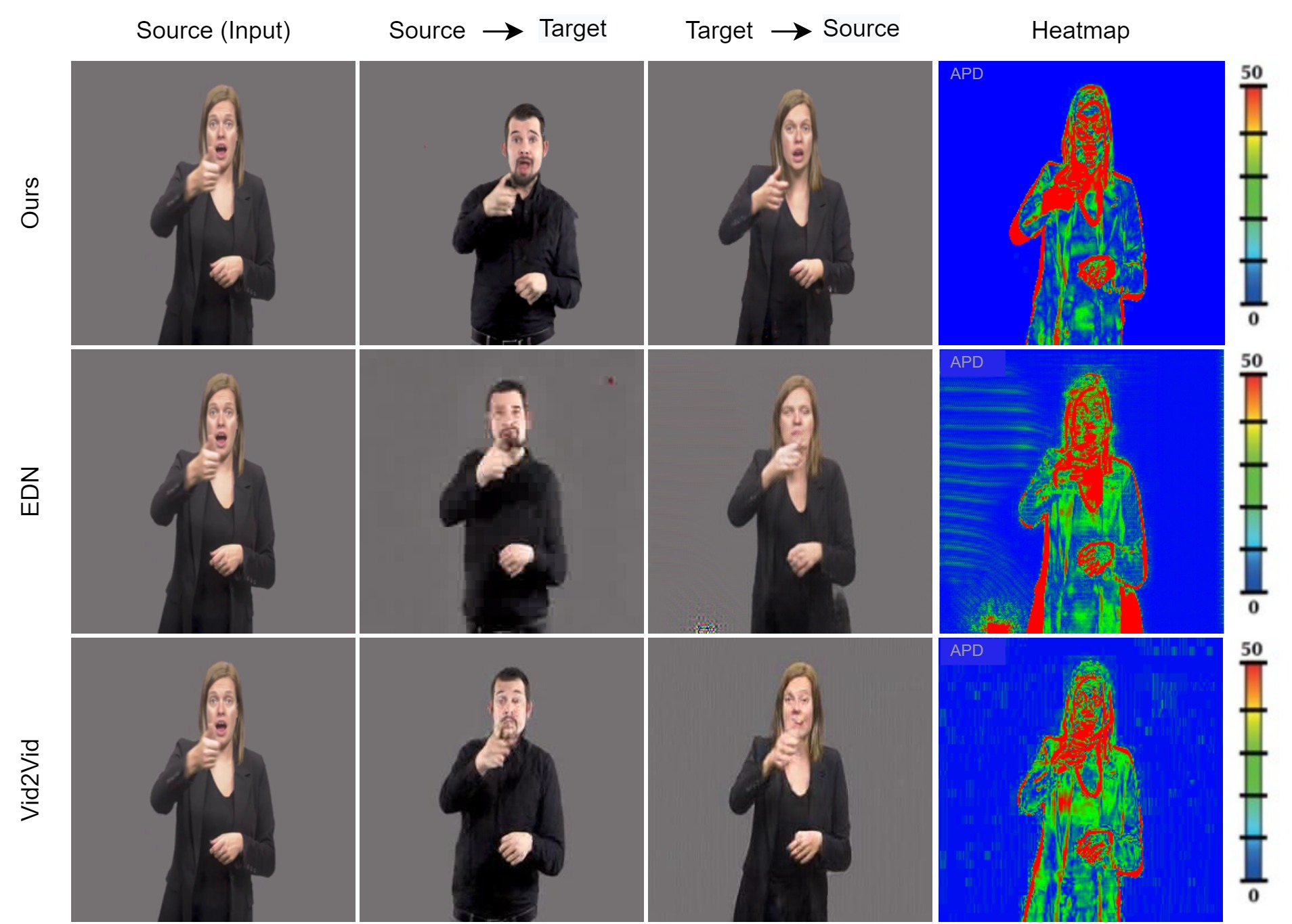}
    \vspace{-0.25cm}
    \caption{Cycle reenactment comparisons with EDN \cite{chan2019everybody} and Vid2Vid \cite{wang2018video}. From left to right: source actor, intermediate-target actor, original source actor driven by the manipulated target actor in the column before, and per-pixel differences between the first and third column.}
    \vspace{-0.4cm}
    \label{fig:cycle_cropped}
\end{figure}

\noindent As reported in the main paper, our method performs better in preserving the source signer's facial expressions and handshapes without distorting the characteristics of the specific identity. Lastly, in Fig. \ref{fig:cycle_cropped}, we extend the qualitative results of cycle reenactment presented in the main paper by providing more comparisons of the various approaches in the $Female \rightarrow Male \rightarrow Female$ cycle reenactment experiment.

\section{Acknowledgments} A. Roussos was supported by the Greek Secretariat for Research and Innovation and the EU, Project SignGuide: Automated Museum Guidance using Sign Language T2EDK-00982 within the framework of “Competitiveness, Entrepreneurship and Innovation” (EPAnEK) Operational Programme 2014-2020. A. Roussos acknowledges also the support by an NVIDIA Academic Hardware Grant Program, which was beneficial in developing and testing the neural rendering models introduced in this paper.

\end{document}